\def\paperlanguage{}
\pdfoutput=1 

\documentclass[letterpaper, 10 pt, conference]{ieeeconf}  

\usepackage{bm}
\usepackage{cite}
\usepackage{flushend}
\include{preamble}

\IEEEoverridecommandlockouts                              

\title{\LARGE \textbf
  {
    \switchlanguage%
    {%
    Development of a Five-Fingerd Biomimetic Soft Robotic Hand\\ by 3D Printing the Skin and Skeleton as One Unit
    }%
    {%
      表皮と骨格を一体に3Dプリントし、五指の動作を有するソフトロボットハンドの開発
    }%
  }
}

\author{Kazuhiro Miyama$^{1}$, Kento Kawaharazuka$^{1}$, Kei Okada$^{1}$, and Masayuki Inaba$^{1}$
  \thanks{$^{1}$ The authors are with the Department of Mechano-Informatics, Graduate School of Information Science and Technology, The University of Tokyo, 7-3-1 Hongo, Bunkyo-ku, Tokyo, 113-8656, Japan.
    {\texttt\small [miyama, kawaharazuka, okada, inaba]@jsk.t.u-tokyo.ac.jp}
  }
}
\begin{document}

\maketitle
\thispagestyle{empty}
\pagestyle{empty}

\begin{abstract}
  \switchlanguage%
  {%
  Robot hands that imitate the shape of the human body have been actively studied, and various materials and mechanisms have been proposed to imitate the human body.
  Although the use of soft materials is advantageous in that it can imitate the characteristics of the human body's epidermis, it increases the number of parts and makes assembly difficult in order to perform complex movements.
  In this study, we propose a skin-skeleton integrated robot hand that has 15 degrees of freedom and consists of four parts.
  The developed robotic hand is mostly composed of a single flexible part produced by a 3D printer, and while it can be easily assembled, it can perform adduction, flexion, and opposition of the thumb, as well as flexion of four fingers.

  }%
  {%
    人体の形状を模倣したロボットハンドはこれまで盛んに研究が行われており、様々な材料や機構での模倣が提案されていた。
    ソフトマテリアルを用いることで人体の表皮の特徴を模倣できるという点で有利であるものの、複雑な動作を行うためには部品点数が増大し組み立ても困難になる。
    本研究では15自由度を有しながら4つのパーツによって構成されている表皮-骨格一体型ロボットハンドを提案する。
    開発したロボットハンドは大部分を3Dプリンタで制作された1つのパーツで構成されており、簡易な組み立てが行える一方で母指の内転、屈曲、対立動作に加え4指の屈曲動作を可能にしている。
    人間が実現可能な16種類の把持姿勢についての実験を行い、16種全てを達成したことを確認した。
  }%
\end{abstract}

\section{Introduction}\label{sec:introduction}
\switchlanguage%
{%
Robots are being used to automate tasks previously performed by humans, with robot hands playing a particularly important role. 
In a social implementation, changing hands according to the task is problematic in terms of implementation cost. 
However, a robot hand that can perform many tasks with a single hand has advantages such as greatly reducing the cost of introduction and contributing greatly to the realization of an automated society. Most tools in society are made to fit human hands, so the human mimetic robot hand can be implemented in society without the use of special tools.\par

Among the human mimetic robot hands, the soft robot hand is particularly effective. Because the organs of the human body, such as the epidermis and tendons, are composed of flexible materials similar to those of the soft robotic hand, the human body can be imitated more faithfully than with a rigid robotic hand. 
According to Shimago et al.'s study\cite{220069}, the use of soft materials in robotic hands has been shown to have three useful properties: a) impact force attenuation, b) conformability, and c) repetitive strain.
These human mimetic robotic hands have been studied for applications such as the reproduction of human tasks\cite{firth2022anthropomorphic} as well as for the study of human anatomy\cite{tian2021design}\cite{7993072}.\par

Soft-material human mimetic robot hands have advantages in terms of hardware and ease of assembly.
In general, rigid robot hands have a large number of parts, and human body-mimetic robot hands in particular are highly complex because they require multiple degrees of freedom. For example, the ShadowHand\cite{shadowhand} has 40 actuators and 20 DoF, making it difficult to maintain. 
On the other hand, the soft robot hand does not need to attach axes and bearings by deforming each axis with flexibility, and can even use bellows\cite{0278364915592961} etc., thus reducing the number of parts.\par

However, there is a problem in that the number of joints and controllability decreases due to the simple design with fewer parts.
Methods to realize other degrees of freedom with a soft robot hand include covering the exterior of a rigid robot hand with soft materials\cite{9762208}\cite{weiner2019embedded} or designing a chamber inside and having it driven by pneumatic pressure\cite{wang2021novel}\cite{9682546}.
But the former method does not reduce the number of parts, and the latter method involves a complex process of layering several layers of silicon to prevent air leakage.\par

In order to solve the problem mentioned above while taking advantage of the soft robot hand, the objective of this study is to develop a soft robot hand that combines the complexity of the model with the simplicity of assembly.\par
We propose a skin-skeleton integrated robot hand(\figref{overview}) that is driven by wires and molded by an optical 3D printer using the same material for the skin and the skeleton to achieve this objective.
\begin{figure}[t]
  \centering
  \includegraphics[width=1.0\columnwidth]{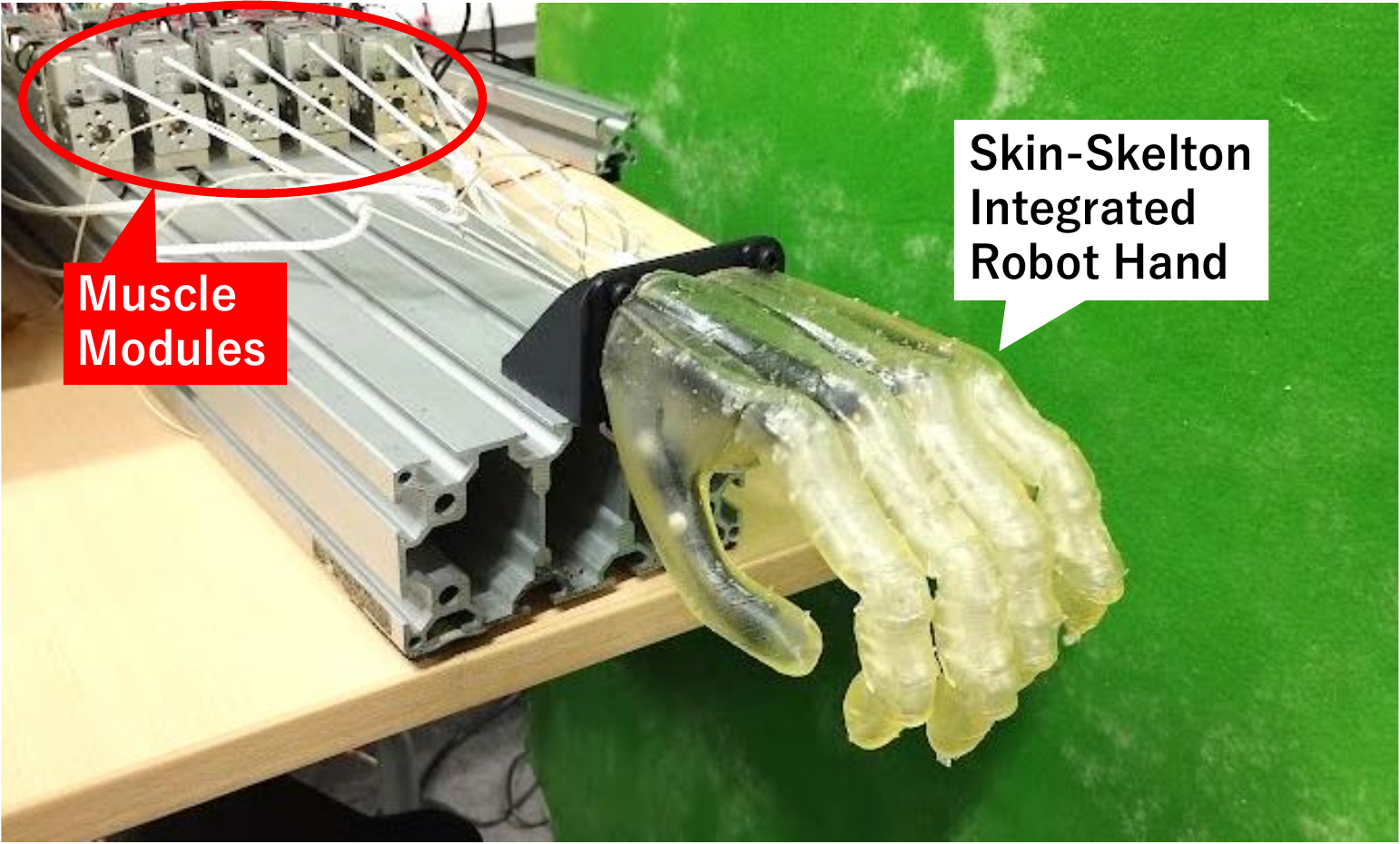}
  \vspace{-3.0ex}
  \caption{Overview of skin-skeleton integratedrobot hand}
  \label{overview}
  \vspace{-3.0ex}
\end{figure}

Robot hands using 3D printers have been studied extensively in the field of soft robotics in recent years. There are methods such as a robot hand that is driven by air pressure by creating a cavity\cite{she2015design} or a hand that is shaped as a single piece\cite{de20223d}, but from a practical standpoint, the low rigidity, the small number of grasping postures that can be realized, and the complexity of the air-driven hand are issues to be solved.
There is research on the use of FDM 3D printers to make the skin and skeleton of the hand itself as a single piece\cite{mohammadi2020practical}, but the FDM method cannot create a hollow space in the part, which weakens the advantage of Shimago's c)repetitive strain dissipation.
The skin-skeleton integrated robot hand is a light fabrication method, which has a skeleton shape and hollow space inside, and is wire-driven, so we thought that both rigidity and structural simplicity could be achieved.

This robot hand has the following three features.  
First, the majority of the hand is composed of a single part, the skin-skeleton integral part, and the entire robot hand is composed of four parts. This simplifies assembly and fabrication, and all parts can be completed simply by sending the model data to a 3D printer. The assembly method is also simple: wires are threaded through the skeleton-skin parts and combined with the metacarpal and thumb parts, and the robot can be completed in less than an hour. \par
Second, the item grasping performance is high. Since the skin-skeleton integrated robot hand has the characteristics of a soft robot hand, it can manipulate a variety of items by grasping utilizing friction and deformation. In addition, since it is a multi-degree-of-freedom under-driven robot hand with five fingers driven by wires, various postures can be realized, which also contributes to improved grasping performance.\par
The third is shock resistance. Because it has soft joints, it can return to its initial posture immediately, although it is greatly deformed when an impact is applied. Therefore, it has high adaptability to the environment and has the potential to be used for cooperation and interaction with humans.\par

In \secref{sec:design}, we describe the outline and overall design of the skin-skeleton integrated robot hand. 
In \secref{sec:skin}, we explain the design of the joints and the palm of the skin-skeleton integrated parts that will be output by the stereolithography 3D printer and verify the effect by experiments.
In \secref{sec:thumb}, the remaining components of the skin-skeleton integrated robot hand, the metacarpal and thumb, are described and their assembly is explained.
In \secref{sec:experiments}, several experiments including grasping experiments using the skin-skeleton integrated robotic hand are conducted to verify its effectiveness as a robotic hand.
In \secref{sec:conclusion}, we conclude the performance of the skin-skeleton integrated robot hand based on the experimental results.
}%
{%
ロボットによる人間が行うタスクの自動化が進められているが、特に重要な役割を担っているのはロボットハンドである。タスクに応じてハンドを変更するのではなく、多くのタスクを一つのロボットハンドで対応できる場合は導入コストが大きく低下し、自動化社会の実現に大きく寄与する。\\
その中でソフトロボットハンドは、剛体ロボットが傷つけてしまう物体を安全に扱うこと,そして様々な形状に合わせて変形することが出来る汎用性の高さなどから様々な研究が行われている。また、Shimago らによる研究\cite{220069}によると、ロボットハンドにソフトマテリアルを用いることによって、 a) impact force attenuation、 b) conformability and c) repetitive strain dissipationの３つの有用な特徴がある。これらの特徴は物体の把持の他、道具の操作などにも活用されている。\par
また、ソフトロボットハンドの応用先として、人体を模倣したハンドがある。人体の表皮や腱と言った器官はソフトロボットハンドと同様に柔軟なマテリアルで構成されているため、剛体ロボットハンドに比べてより忠実に人体を再現することができる。こういった人体模倣ロボットハンドは人間が行うタスクの再現\cite{firth2022anthropomorphic}の他、人体構造の考察\cite{tian2021design}\cite{7993072}と言った用途で研究が行われている。\\
ソフトマテリアルによる人体模倣ロボットハンドは、ソフトウェアの面で制御負荷の少ないシステムを構築できる他\cite{lee2017soft}、ハードウェアの面でも組み立て性に利点がある。一般に剛体ロボットハンドは部品点数が多く、特に人体模倣ロボットハンドは多自由度が求められることから複雑性が高い。例えばShadowHand\cite{shadowhand}は40個のアクチュエータと20DoFを有しており、メンテナンスの難易度が高い。一方でソフトロボットハンドは、各軸を柔軟性で変形させることにより軸や軸受を取り付ける必要がなく、蛇腹などを用いることができる\cite{0278364915592961}ため、部品点数を少なくすることが可能になる。\\
ただ、部品点数が少なくシンプルな設計となることによって関節数や制御性が減少する問題がある。ソフトロボットハンドで他自由度を実現するための手法として、剛体ロボットハンドの外装をソフトマテリアルで覆う手法\cite{9762208}\cite{weiner2019embedded}や、内部にチャンバを設計して空気圧で駆動させる方式などがある\cite{wang2021novel}\cite{9682546}。しかし、前者は部品点数のは減らせておらず、後者はエアの漏れを防ぐためにシリコンを何層にも重ねる複雑な工程を経て製作している。\par

本研究では、ソフトロボットハンドの利点を活かしながら前述の問題を解決するため、モデルの複雑さと組み立ての簡易性を両立するソフトロボットハンドを開発することを目的とした。\\
本目的を達成させるために、表皮と骨格を同一の素材で光造形式3Dプリンタによる成形を行い、ワイヤによって駆動する表皮骨格一体型ロボットハンド(\figref{overview})を提案する. 3Dプリンタを用いたロボットハンドは、ソフトロボットの分野で近年多く研究が成されている。空洞を作って空気圧によって駆動させるロボットハンド\cite{she2015design}や、ハンドそのものを一体で整形する方式\cite{de20223d}はあるものの、実用面では剛性の低さや実現可能な把持姿勢の少なさ、エア駆動であることによる複雑性などが課題になっている。表皮骨格一体型ロボットハンドは内部に骨格形状を有しワイヤ駆動であるため、剛性と構造の簡易性を両立することが出来ると考えた。
本ロボットハンドは以下の3つの特徴を持つ。

\begin{figure}[b]
  \centering
  \includegraphics[width=1.0\columnwidth]{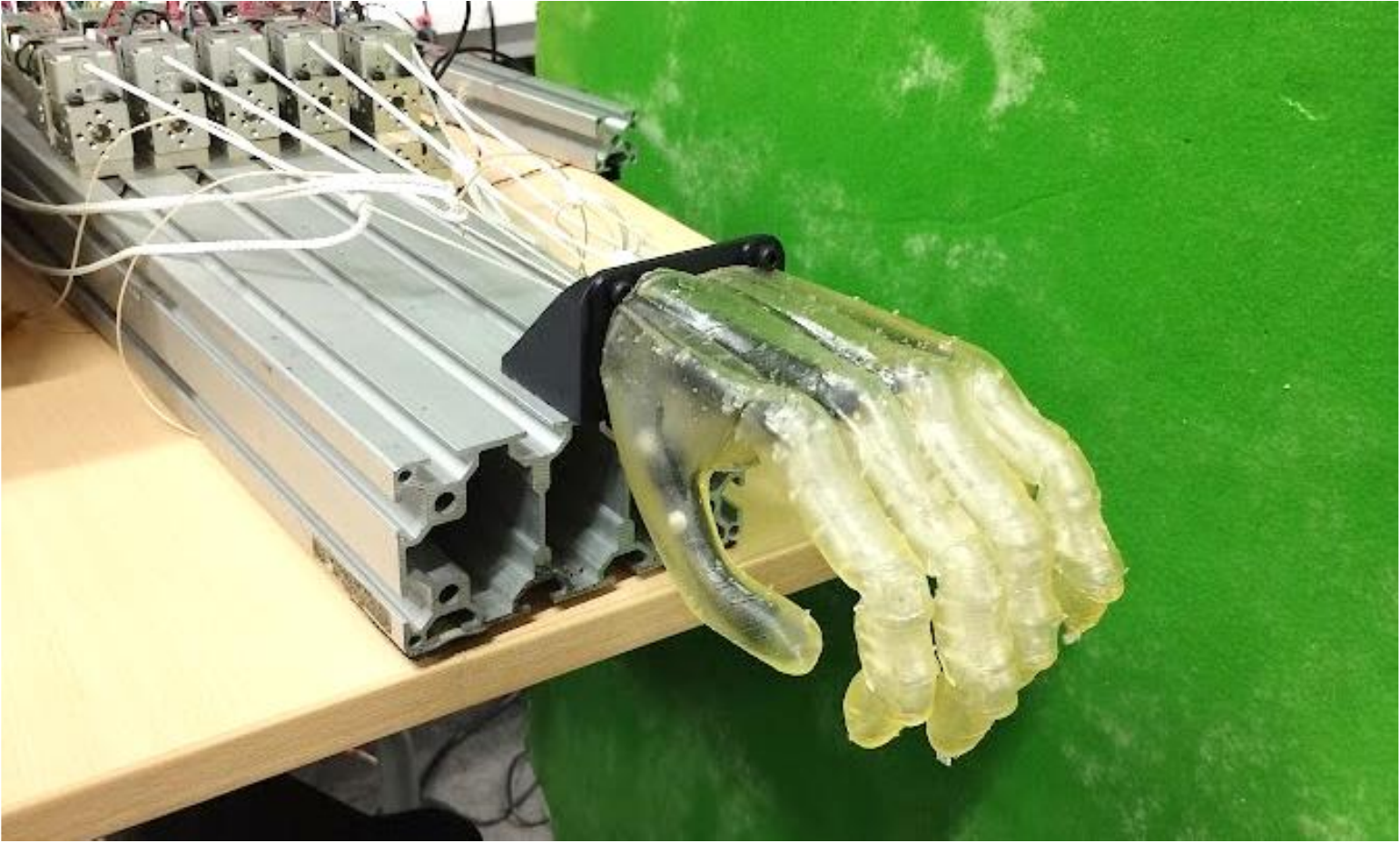}
  \caption{Overview of skin-skeleton integratedrobot hand}
  \label{overview}
  \end{figure}

第一に、ハンドの大部分が表皮-骨格一体パーツという一つの部品で構成されている上に、ロボットハンド全体を構成しているのは4つパーツである点である。これによって組み立てや製作が簡易であり、実際に3Dプリンタにモデルデータを送信するだけで全てのパーツは完成する。また、組立方法も製作した骨格ー表皮パーツにワイヤを通して中手骨パーツと親指パーツと組み合わせるのみであり、1時間足らずで完成させることができる。\par
第二に、アイテム把持性能が高い。骨格-表皮一体ロボットハンドはソフトロボットハンドの特徴を有するため、摩擦や変形を活用した把持で多様なアイテムの操作が可能になる。それに加えて、五指がワイヤによって駆動する多自由度劣駆動ロボットハンドであるため様々な姿勢を実現することが可能であり、これも把持性能の向上に寄与している。\par
第三に、耐衝撃性である。柔らかい関節を有しているため、衝撃を加えた際に大きく変形するものの、すぐに初期姿勢に戻ることができる。そのため高い環境適応性を有しており、人間との協働やインタラクションに活用できる可能性がある。\par

\secref{sec:design}では表皮骨格一体型ロボットハンドの概要と、全体の設計について述べる。 
\secref{sec:skin}では光造形式3Dプリンタで出力する表皮骨格パーツの関節部、掌の設計について説明し、実験によって効果を検証する。
\secref{sec:thumb}では表皮骨格一体型ロボットハンドを構成する残りの要素である中手骨と親指について述べ、組み立てに関する説明を行う。
\secref{sec:experiments}では表皮骨格一体型ロボットハンドを用いた把持実験を始めとする複数の実験を行い、ロボットハンドとしての効果を検証する。
\secref{sec:conclusion}では実験結果を踏まえて表皮骨格一体型ロボットハンドのハンドとしての性能について結論を述べる.
}%

\section{Overview of the skin skeleton-integrated robot hand} \label{sec:design}
\switchlanguage%
{%
The overall view of the skin-skeleton integrated robot hand is shown in \figref{overview}. This robot hand has a five-finger skeletal structure that mimics the human hand and is driven by seven wires (hereinafter referred to as tendons).
The index, middle, ring, and little finger, excluding the thumb, have one tendon for each finger, and the flexion motion is performed by pulling the tendon. The extension motion of these fingers is performed by the elasticity of the joints.
The thumb is driven by three wires, and the placement of the wires and other details are described in detail in \secref{subsec:comparison}. \\
The robot hand is composed of a skin-skeleton integrated part, a metacarpal part, and a thumb part.
The skin-skeleton integrated part is made of a Shore hardness 50A soft material (Elastic 50A, RS-FL-ENG-E50A) and is fabricated using a light-molded 3D printer (SLA, Stereo Lithography Apparatus) Form3. On the other hand, the metacarpal and thumb are made of ABS filament.
The overall dimensions of the robot hand are based on the AIRC hand size data\cite{handsize}, and the diameter and joint lengths are set to be similar to those of an adult male.
}%
{%
表皮骨格一体型ロボットハンドの全体像を\figref{overview}に示す。本ロボットハンドは人体を模倣した五指骨格構造であり、7本のワイヤ(以下、腱とする)によって駆動する。母指を除くindex, middle, ring, little fingerは各指に一本の腱が配置されており、それを引くことによって屈曲動作を行う。なお、それらの伸展側の動作は関節の弾性力を利用している。母指は3本のワイヤによって駆動する。ワイヤの配置などは\secref{subsec:comparison}で詳細を述べる。\\
ロボットハンドは表皮パーツ、中手骨パーツ、親指パーツの3つで構成されており、表皮パーツはショア硬度50Aのソフトマテリアル(Elastic 50A、RS-FL-ENG-E50A)を用い光造形3Dプリンタ(SLA, Stereo Lithography Apparatus)のForm3で製作されている。一方で中手骨と親指はABS樹脂で制作されている。また、ロボットハンド全体の寸法はAIRCの手の寸法データ\cite{handsize}を元に、成人男性のものと類似するように直径や関節長を設定した。\\
}

\subsection{Comparison with the human hand}\label{subsec:comparison}
\begin{figure}[htb]
  \centering
  \includegraphics[width=1.0\columnwidth]{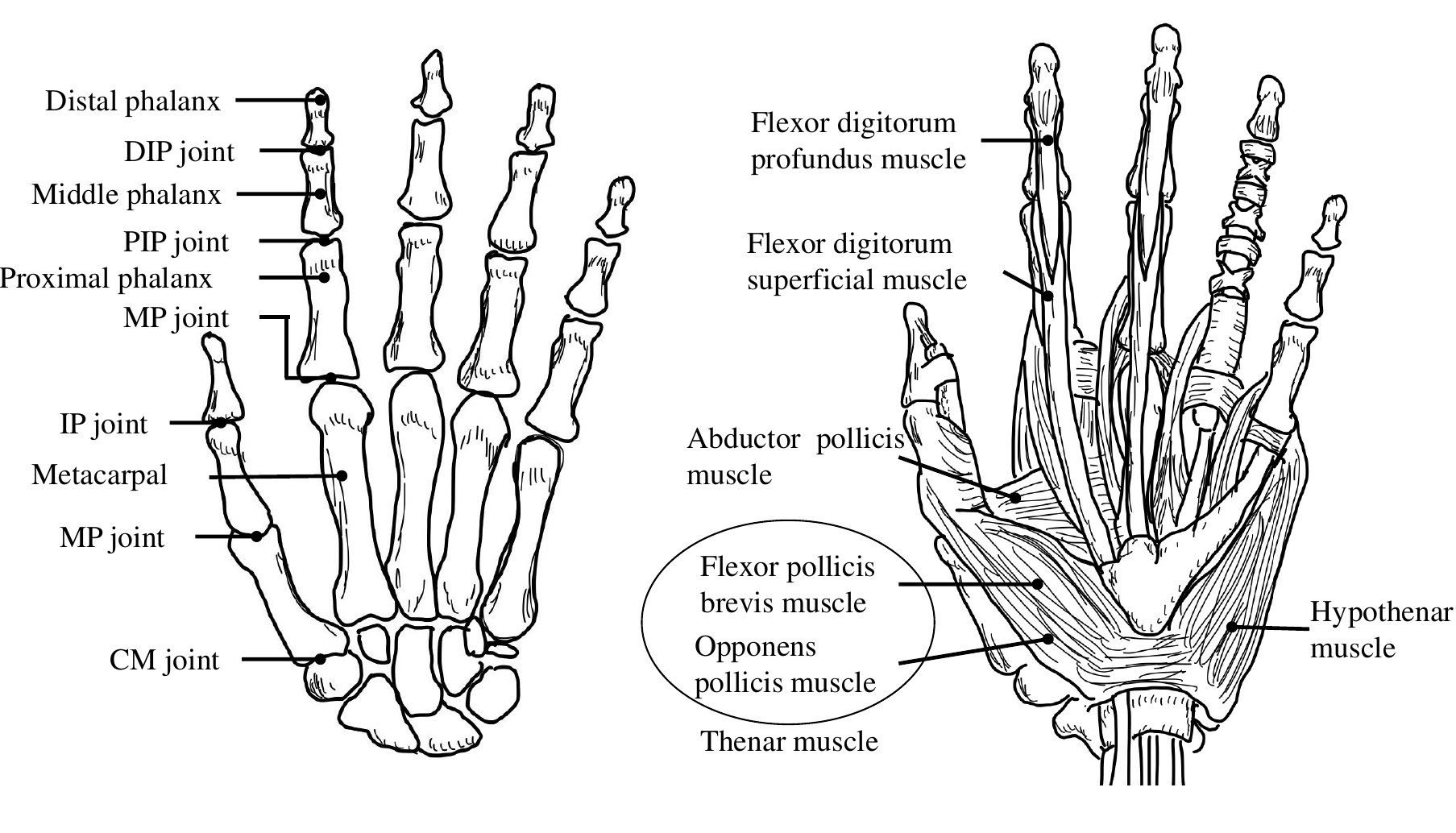}
  \vspace{-3.0ex}
  \caption{Structure of the human hand and names of its parts}
  \label{human-hand}
  \vspace{-3.0ex}
  \end{figure}

\switchlanguage%
{%

  We will compare this robot hand with a human hand (\figref{human-hand}). First, we discuss the four fingers, excluding the thumb.
  They have three joints: the DIP joint, the PIP joint, and the MP joint, which are made up of four bones: the distal phalanx, middle phalanx, proximal phalanx, and metacarpal bones. 
  The flexor digitorum profundus, flexor digitorum teres, and interosseous muscles are allocated to each of these joints, and the DIP, PIP, and MP joints are driven by the flexor digitorum profundus, and the PIP and MP joints are driven by the flexor digitorum superficialis. This robot hand reproduces the flexor digitorum profundus that drives the three joints simultaneously and realizes flexion movements by placing one tendon on each finger.
  Although human fingers also have dorsal interosseous muscles for adduction and abduction, these muscles are not used for grasping and tool manipulation, so they were not employed in this study. \par

  The thumb in the human body is similar to the other four fingers in that it has an IP joint and MP joint with the distal phalanx, proximal phalanx, and metacarpal bones, but it is unique in that the CM joint, the joint where the metacarpal bone is connected to the wrist side, has two degrees of freedom. In other words, the complexity of the thumb is established by the flexion-extension of the MP joint as well as the adduction-abduction and opposition-resumption of the CM joint.
  The flexor pollicis muscle and extensor pollicis muscle are responsible for flexion and extension, the adductor pollicis muscles and abductor pollicis muscles are responsible for adduction and abduction, and the opponens pollicis muscle and retractor pollicis longus muscles are responsible for opposition and return.
  In this robot hand, the flexor pollicis brevis muscle, aductor pollicis muscles, and opponens pollicis muscle are used to drive the thumb.
  This robot hand has the same degrees of freedom as the human body and is driven by three tendons. Details on the drive of the thumb will be described in detail in \secref{sec:thumb}.\par
  The functions of the palms are also compared. The palm can be broadly divided into four parts by the MP joints, the thenar muscles, and the hypothenar muscles.
  By moving the MP joints of each finger, the shape of the palm can be changed to assume various postures, such as holding or pinching an object. The details will be described in \secref{subsec:palm}, but for this robot hand, we also designed the palm for flexion of the MP joints and movement of the CM joints of the thumb.\par
  The human body is said to have 20 DoFs, and this robot hand has 15 DoFs, 3 DoFs for each of 5 fingers.

}%
{%
 次に、本ロボットハンドが模倣した人間の手(\figref{human-hand})との比較を行う。まず、母指を除く4指について述べる。これらは末節骨、中節骨、基節骨、中手骨の4つの骨によるDIP関節、PIP関節、MP関節の3関節を持つ。それぞれには深指屈筋、浅指屈筋、骨間筋、虫様筋などが配置されており、深指屈筋によってDIP,PIP,MP関節が、浅指屈筋によってPIP、MP関節が駆動する。本ロボットハンドでは3関節を同時に駆動させる深指屈筋を再現し、各指に1本の腱を配置することで屈曲動作を実現している。人間の指にはこれ以外にも内転、外転動作を行うための背側骨間筋などがあるが、今回行う把持や道具操作などでは用いないため採用しなかった。\par
 人体における母指は、末節骨、基節骨、中手骨の3つによるIP関節、MP関節を持つことは他の4指と類似しているが、中手骨が手首側に接続される関節であるCM関節が2自由度を持つことが特徴的である。つまり、MP関節の屈曲-伸展に加えてCM関節の内転-外転、対立-復位によって親指の複雑性が成立していると言える。屈曲-伸展は母指屈筋、母指伸筋によって、内転-外転は母指内転筋、母指外転筋によって、対立-復位は母指対立筋、長母指伸筋によって成りたっている。本ロボットハンドでは親指を駆動させるための筋肉として、短母指屈筋、母指内転筋、母指対立筋を再現した。これによって本ロボットハンドは人体と同様の自由度を持ち、3本の腱によって駆動する。母指の駆動に関する詳細は\secref{sec:thumb}で詳細を述べる。\par
 また、掌の機能についても比較を行う。掌はMP関節と母指球筋、小指球筋によって4つのパーツに大別することが出来る。各指のMP関節を動かすことで掌の形を変え、物体を固定したり挟み込むような様々な姿勢を取ることができる。詳細は\secref{subsec:palm}で述べるが、本ロボットハンドでもMP関節の屈曲のための掌の設計を行った。\\
 人体は20自由度を持つと言われているが、本ロボットハンドは5指それぞれ3自由度ずつの15自由度を有している。
}%

\section{Design of Skin-Skelton Integrated Part} \label{sec:skin}
\begin{figure}[htb]
  \centering
  \includegraphics[width=0.7\columnwidth]{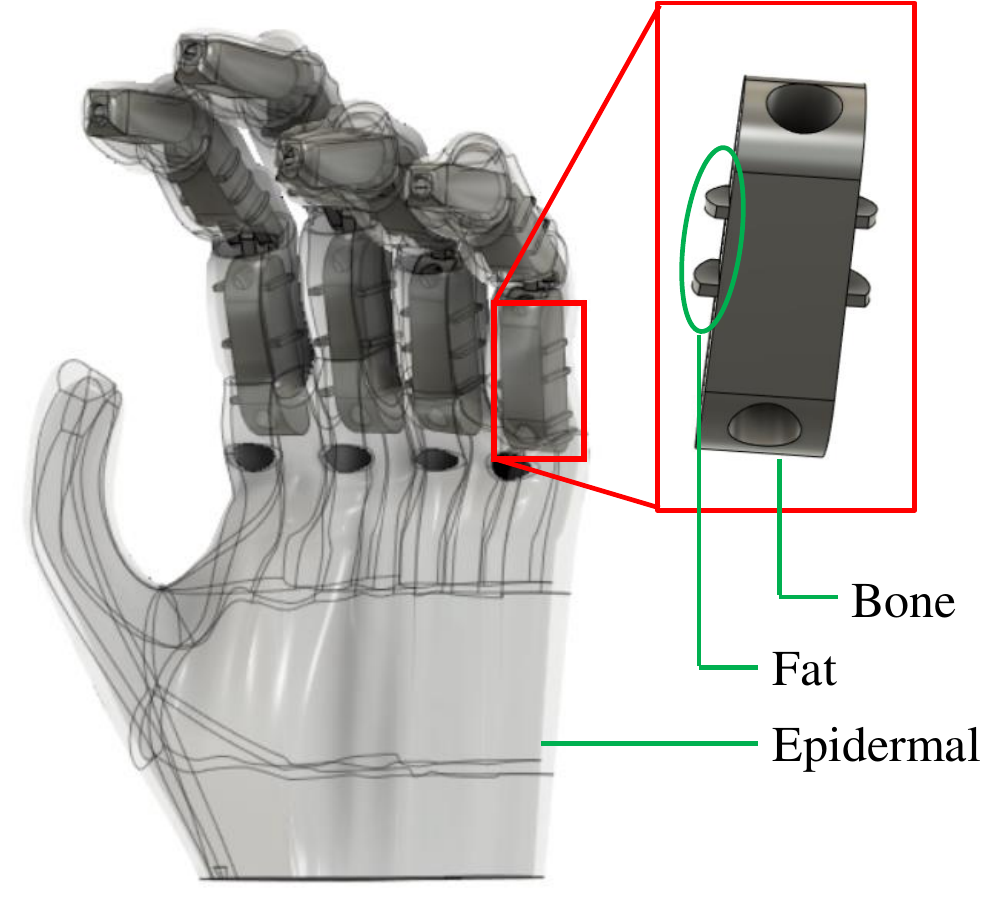}
  \caption{Overview of skin-skeleton integrated parts}
  \label{skin-config}
  \vspace{-3.0ex}
  \end{figure}

\switchlanguage%
{%
This chapter describes the skin-skeleton-integrated part, which is the largest component of the robot hand (\figref{skin-config}).
This part consists of a model that mimics the epidermis of the human body, with the distal phalanx, middle phalanx, and proximal phalanx of the second to fifth fingers. The epidermis and the skeleton are connected by a layer of fat called the fat layer. The fat layer is a plate-like model about 1[mm] thick that extends from the skeleton to the epidermis at equal intervals to anchor the skeletal parts. \par

As mentioned above, this part is produced by SLA 3D printer, which is difficult for commonly used Fused deposition modeling (FDM) 3D printers.
The FDM printer can print filaments such as TPU with similar hardness to the resin used in this study, but it is difficult to model as designed because many supports are required for the framework inside the skin, and the strength in the direction parallel to the stacking direction is weak.
However, the use of SLA printers makes it possible to avoid using supports inside the skeleton and to ensure strength regardless of the stacking direction. \par 
The thickness of the skin is a trade-off between break resistance and flexibility. If it is too thin, it will break due to impact or tension, while if it is too thick, it will break due to folding during flexion, in addition to not providing flexibility. 
As a result of testing several thicknesses, it is confirmed that the thickness of 2[mm] for most parts of the robot hand and 0.5[mm] for parts where flexibility is required, such as joints, is the method that causes the fewest ruptures, and this method is adopted in this study as well.
The following four points are considered important in the design of this robot hand.

\begin{quote}
  \begin{enumerate}
    \item Structure of joints
    \item Design of skeletal and tendon
    \item Initial posture
    \item Palm flexibility
  \end{enumerate}
 \end{quote} 
 The next subsections will discuss these in more detail.
}%
{%
  本章では表皮-骨格一体型ロボットハンドを構成する最大の要素である表皮-骨格一体パーツについて述べる(\figref{skin-config})。本パーツは人体の表皮を模倣したモデルの中に第二指から第五指までの末節骨、中節骨、基節骨を有する構造をしている。表皮と骨格は脂肪層と呼ばれるパーツによって結合されている。脂肪層は厚さ1[mm]ほどの板状のモデルで、骨格から等間隔に表皮に伸びて繋がることで骨格部の固定を行っている。\par
  前述のように本パーツはSLA3Dプリンタによって制作されており、一般に用いられている熱融解積層型(Fused deposition modeling, FDM)3Dプリンタでは困難である。FDMプリンタではTPUなどのフィラメントは今回使用したレジンと同様の硬度を持つものは印刷可能であるものの、表皮内部の骨格に多数のサポートが必要になることから設計通りの造形が困難であり、積層方向に平行な方向の強度が弱くなる問題があった。しかし、SLAプリンタを用いることで、骨格内部にサポートを用いないこと、積層方向によらない強度を確保することが可能になっている。\par
  表皮の厚みは耐破断性と柔軟性のトレードオフとなる。薄すぎる場合は衝撃や引張によって破断してしまう一方で、厚すぎる場合は柔軟性を確保できない他にも屈曲の際に折込まれることで破断が発生する。いくつかの厚みで検証をした結果、ロボットハンドの多くの箇所で2[mm]とし、関節部など柔軟性が求められる箇所は0.5[mm]とする方式が最も破断が少ないことが確認されたため本研究でもこれを採用した。\par
  本ロボットハンドの設計において重要視した箇所は以下の4点である。

\begin{quote}
  \begin{enumerate}
   \item 関節部の構造
   \item 骨格と腱の配置
   \item 初期姿勢
   \item 掌の柔軟性
  \end{enumerate}
 \end{quote} 
 次節以降ではこれらの詳細について述べる。
}%

\subsection{Structure of the Joint of Skin-Skelton Integrated Parts} \label{subsec:joint}
\begin{figure}[htb]
  \centering
  \includegraphics[width=0.8\columnwidth]{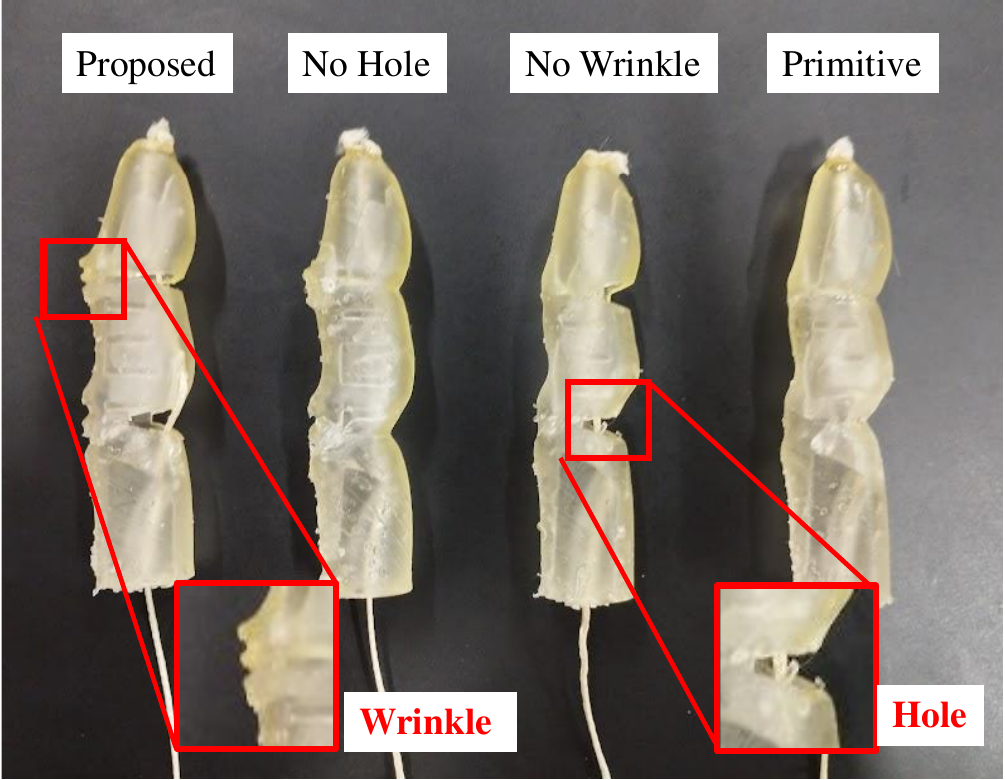}
  \caption{Finger models used in tension tests}
  \label{fingers}
  \end{figure}

\begin{figure}[htb]
  \centering
  \includegraphics[width=0.8\columnwidth]{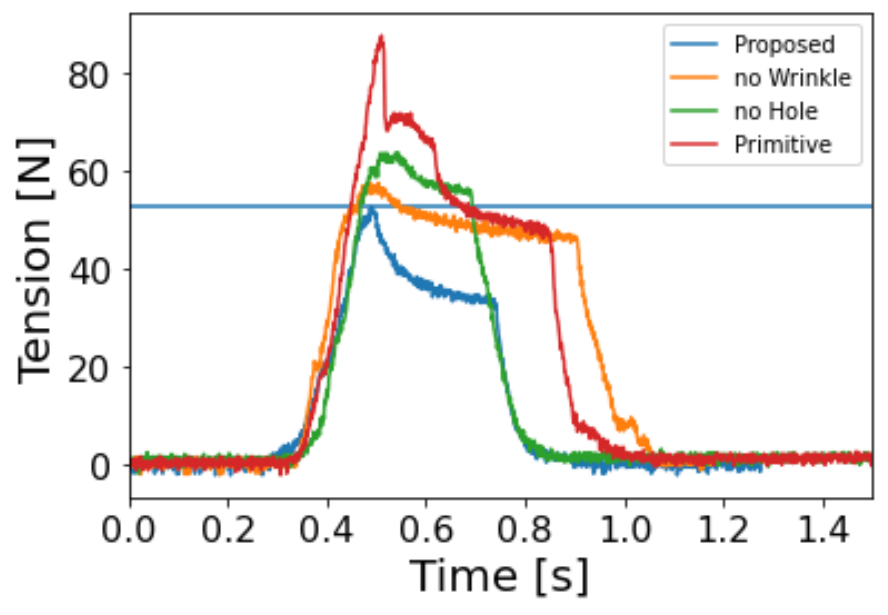}
  \caption{Resulting graph of tension due to flexion of each finger model.(The blue line is the maximum tension of the proposed method)}
  \label{ex-hand-result}
  \end{figure}

\switchlanguage%
{%
First, the joint structures of the four fingers, excluding the thumb, are described. The material used in the fabrication of this robot hand was a flexible material with a Shore hardness of 50A and an elongation at break of 160\%.
 However, since the breaking tensor was as low as 3.2 MPa, it was confirmed that the material would break under pressure, such as when folded or pulled. Therefore, holes are designed on the palm side of the four finger joints (the same at the PIP, DIP, and MP joints), and a structure with exaggerated human wrinkles on the dorsal side (only at the PIP and DIP joints).
 The holes on the palm side serve to prevent folding and breaking during flexion, while the wrinkles on the dorsal skin are designed to prevent tearing through extension.\par
 In addition to this, the skin was designed to be about 0.5[mm] thick around the joints, whereas the thickness of the surrounding skin is about 2[mm], to encourage more bending. In order to promote deformation due to pressure, we also design the palm side of the hand to create a layer of air.
 This resulted in a system that increases the contact area upon contact with an object on the human body, with the epidermal part deforming until it hits the skeletal part.\par

 Experiments were conducted to verify the effect of the additional design of dorsal-side wrinkles and palm-side holes in the finger joints: 1) a model of the proposed method with holes on the palm side and wrinkles on the dorsal side, 2) a model with holes on the palm side and a flat dorsal side, 3) a model without holes on the palm side and wrinkles on the dorsal side, 4) a model without holes on the palm side and wrinkles on the dorsal side.
 The models of each of the four types of fingers were created (\figref{fingers}) and the tension in flexion was measured with the attached load cell while fixed to the muscle module \cite{asano2015sensordriver}. The resulting graphs and maximum tension are shown in \figref{ex-hand-result} and \tabref{tab:ex-hand-result}. \par
 The blue line in \figref{ex-hand-result} is the value of the maximum tension of the proposed method. From these results, there was no model that required less tension for bending than the proposed method. 2) and 3) Both methods were found to require less tension for bending when there was a slight wrinkle, although there was no significant difference.
 However, it was found that the absence of either element required 166\% more tension to perform the flexion action than the presence of either element. These results confirm that the dorsal-side wrinkle and the palm-side hole contribute to the reduction of tension for the flexion movement.

}%
{%
  まず、母指を除く4指の関節構造について述べる。本ロボットハンドの製作で用いられた素材はショア硬度50Aという柔軟性を持ち、破断伸度が160\%という伸長しやすい素材であった。しかしながら、破断テンソルが3.2[MPa]と小さいため、折り畳まれたり引っ張られたりなどの圧力がかかると破断することが確認された。そのため、4指の関節の掌側に穴を(PIP、DIP、MP関節で同様)、背側に人間のシワを誇張した構造を(PIP,DIP関節のみ)設計した。手の平側の穴は屈曲時に折り畳まれて破断することを防ぐ役割があり、背皮のシワは伸展によって裂けることを防ぐために設計されている。\\
  これに加え、周辺の皮膚の厚さは約2[mm]なのに対し関節周辺では0.5[mm]程度に設計することで，より曲がりやすさを促した。また、加圧による変形を促すため、手の平側に空気の層を作るような設計も行った。これによって、骨格部に当たるまで表皮部が変形するという、人体にある物体接触時に接触面積が増大するシステムとなった。\\
  指関節における背側のシワと掌側の穴の追加設計の効果検証の実験を行った。1)手の平側に穴があり背側にシワがある提案手法のモデル、2)手の平側に穴があり背側が平坦なもの、3)手の平側に穴はなく背側にシワがあるもの、4)手の平側に穴がなく背側にもシワがないもの の4種類の指のモデルをそれぞれ作成し(\figref{fingers})、筋モジュール\cite{asano2015sensordriver}に固定した状態で、付属のロードセルによって屈曲時の張力を測定した。結果のグラフと最大張力を\figref{ex-hand-result}と\tabref{tab:ex-hand-result}に示す。\par
  \figref{ex-hand-result}の青線が提案手法の最大張力の値である。これらの結果より、提案手法をよりも屈曲に必要な張力が小さいモデルは存在しなかった。2)、3)ともに大きな違いはないものの、わずかにシワがある方が屈曲に必要な張力が小さいことがわかった。しかしながら、どちらの要素もない場合はある場合と比べ屈曲動作を行うために166\%の張力が必要になることがわかった。以上のことから、背側のシワと手の平側の穴が屈曲動作の張力減少に寄与していることを確認できた。
}%

\begin{table}[htbp]
  \caption{Characteristics of the fingers used in the experiment and the maximum tension}
  \label{tab:ex-hand-result}
  \centering
  \begin{tabular}{c|cc}
    & Hole on palm side & No Hole \\
      \hline
      Wrinkle on dorsal side &52.7[N] & 63.9[N]\\
      No Wrinkle & 57.6[N] & 87.8[N]\\
  \end{tabular}
\end{table}

\subsection{Design of the Skelton and Tendon}
\begin{figure}[htb]
  \centering
  \includegraphics[width=1.0\columnwidth]{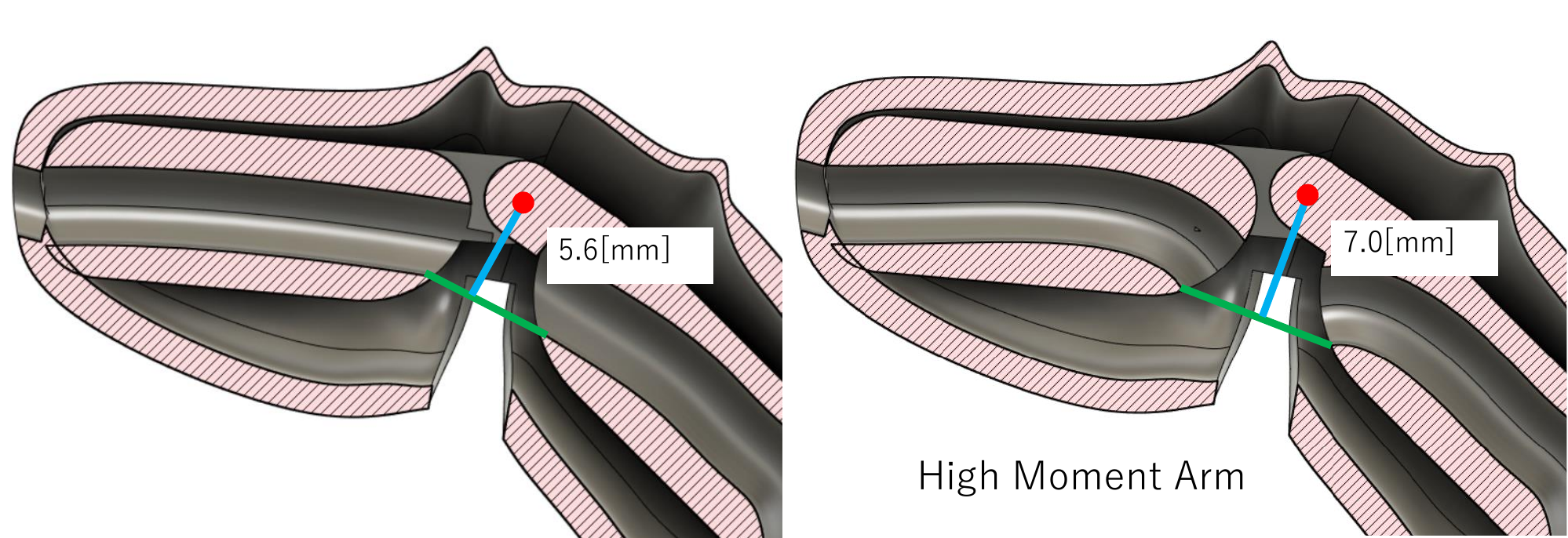}
  \vspace{-3.0ex}
  \caption{Cross section of the index finger and moment arm of the PIP joint}
  \label{cross-skelton}
  \end{figure}

\switchlanguage%
{%
  In this section, the skeletal and tendon structures are described. The size of the skeleton, as well as the entire finger, is based on the AIRC hand dimension data\cite{handsize}, and based on the data, the design was made so that the distal:middle:basal segments are approximately 2:3:5.
  A cross-sectional view of the index finger is shown in \figref{cross-skelton}. The joint is designed as a hinge at the center of rotation, which prevents the finger from shifting in the adduction-abduction direction. \par
  The tendons for drive are placed inside the skeleton. Although tendons in the human body exist on the outside of the skeleton, the problem of high pressure applied to the epidermis by the stretched tendons arose when reproduced in this robot hand.
  In the human body, a thin membrane-like organ called the tendon sheath covers the tendon and drives the tendon while it remains closely attached to the finger bone.
  Although attempts to reproduce this in the robot hand would place high pressure on the thin tendon sheath, two problems arise when the tendon is placed without the tendon sheath. One is that the moment arm becomes too large, resulting in a small joint flexion angle, and the other is that the epidermis is subjected to high pressure from the tendon.
  Since experiments using tendons placed on the outside of the skeleton confirmed rupture of the epidermis, we decided to use a method in which the tendons are placed inside the skeleton. \par
  The hole inside the skeleton as a guide for the tendon is curved to secure the moment arm to drive the tendon. Here, using the DIP joint of the \figref{cross-skelton} as an example, the center of rotation is considered as the red point in the figure.
  When the wire guide is placed at the centerline of the skeleton, the moment arm was 5.6 [mm]. In contrast, by placing the wire at a distance of $a$ from the end point of the skeleton and at an angle of $\theta$ to the center line, the moment arm of $a \tan \theta$ can be increased. The value of a at the DIP joint of the index finger is 3 [mm] and $\theta$ is 30°, indicating an increase in moment arm of 25\%.
  When the wire guide inside the skeleton is aligned with the centerline, it did not flex properly due to the tendon path and moment arm problem, but by properly designing the wire guide inside the skeleton, flexion was possible and the required tension was greatly reduced. The moment arm of the human joint protected by the tendon sheath is 7.5[mm]\cite{neumann2013kinesiology}, which is similar to the 7.0[mm] of this robot hand.
}%
{%
本節では、骨格と腱の構造について述べる。骨格のサイズは指全体と同じくAIRCの手の寸法データ\cite{handsize}を参考にしており、データに基づいて末節：中節：基節が概ね2:3:5となるように設計した。\figref{cross-skelton}に示指の断面図を示す。関節部は回転中心となる部位にヒンジのような設計を施し、これによって指が内転-外転方法にずれることを防いでいる。\par
また、駆動のための腱は骨格の内部に配置している。人体の腱は骨格の外側に存在しているものの、本ロボットハンドで再現した場合は伸び切った腱が表皮に高い圧力が加える問題が生じた。人体では滑膜性腱鞘という薄い膜状の器官が腱を覆っており、指の骨に密接したまま駆動する。これをロボットハンドで再現しようとすると薄い腱鞘に高い圧力が加わってしまうものの、腱鞘無しで腱を配置すると2つの問題が生じる。一つはモーメントアームが大きくなりすぎるため関節の屈曲角が小さくなってしまうという点で、もう一つは表皮に腱による大きな圧力が加わることである。骨格の外側に腱を配置した実験によって表皮の破断を確認したため、骨格の内部に腱を配置する方式とした。\par
腱のガイドとしての骨格内部の穴が湾曲しているのは、腱を駆動させるためのモーメントアームを確保するためである。ここで、\figref{cross-skelton}のDIP関節を例に、回転中心を図中の赤点として考える。骨格の中心線にワイヤガイドを設置した場合、モーメントアームは5.6[mm]だった。これに対し、骨格の端点から$a$の距離で中心線に対し$\theta$傾けた位置にワイヤを出すことで、$a \tan \theta$モーメントアームを増大させることができる。示指のDIP関節におけるaの値は3 [mm], $\theta$は30°であるため、25\%のモーメントアームを増加させていることがわかる。骨格内部のワイヤガイドを中心線に沿わせた場合は腱経路とモーメントアームの問題で正常に屈曲しなかったが、骨格の中のワイヤガイドを適切に設計することで屈曲が可能になった上に必要張力を大きく下げることができた。腱鞘によって保護された人体の関節のモーメントアームは7.5[mm]であり\cite{neumann2013kinesiology}、本ロボットハンドの7.0[mm]と類似している.
}%

\subsection{Initial Posture of the Robot Hand}
\switchlanguage%
{%
The initial posture of this robot hand is slightly flexed, which is designed to mean the balanced position of the muscles on the extension and flexion sides.
The human hand has muscles and tendons on the flexion side and the extension side, and in the natural state when both muscles are relaxed, the hand is in a slightly flexed position. Since this robot hand has an actuator only on the flexion side while the extension side is restored by elasticity, we thought it would be unnatural to design it in the extended state.\par
Therefore, the initial angles of the finger joints, which were confirmed in experiments using cadavers\cite{nimbarte2008finger}, are used as a reference, and the initial postures are 35° for the MP joint, 40° for the PIP joint, and 15° for the DIP joint from the fully extended state. 
This reduced the angular change required for flexion and decreased the tension required for grasping. \par
}%
{%
  本ロボットハンドの初期姿勢はわずかに屈曲した状態だが、これは伸展側と屈曲側の筋肉の釣り合いの位置を意味して設計されている。人間の手には屈曲側と伸展側に筋や腱が配置されており、両者の筋を弛緩させた自然状態ではわずかに屈曲した状態となる。本ロボットハンドは屈曲側のみアクチュエータを有している一方で伸展側は弾性力で復元するため、伸展した状態で設計することは不自然であると考えた。
  そこで、死体を用いた実験で確認された指関節の初期角度\cite{nimbarte2008finger}を参考にし、伸展しきった状態からMP関節では35°、PIP関節では40°、MIP関節では15°傾けるような初期姿勢としている。これによって屈曲に必要な角度変化が減少し、把持の際の必要張力を減少させることができた。\par
}%

\subsection{Design of Palm}\label{subsec:palm}
\begin{figure}[htb]
  \centering
  \includegraphics[width=1.0\columnwidth]{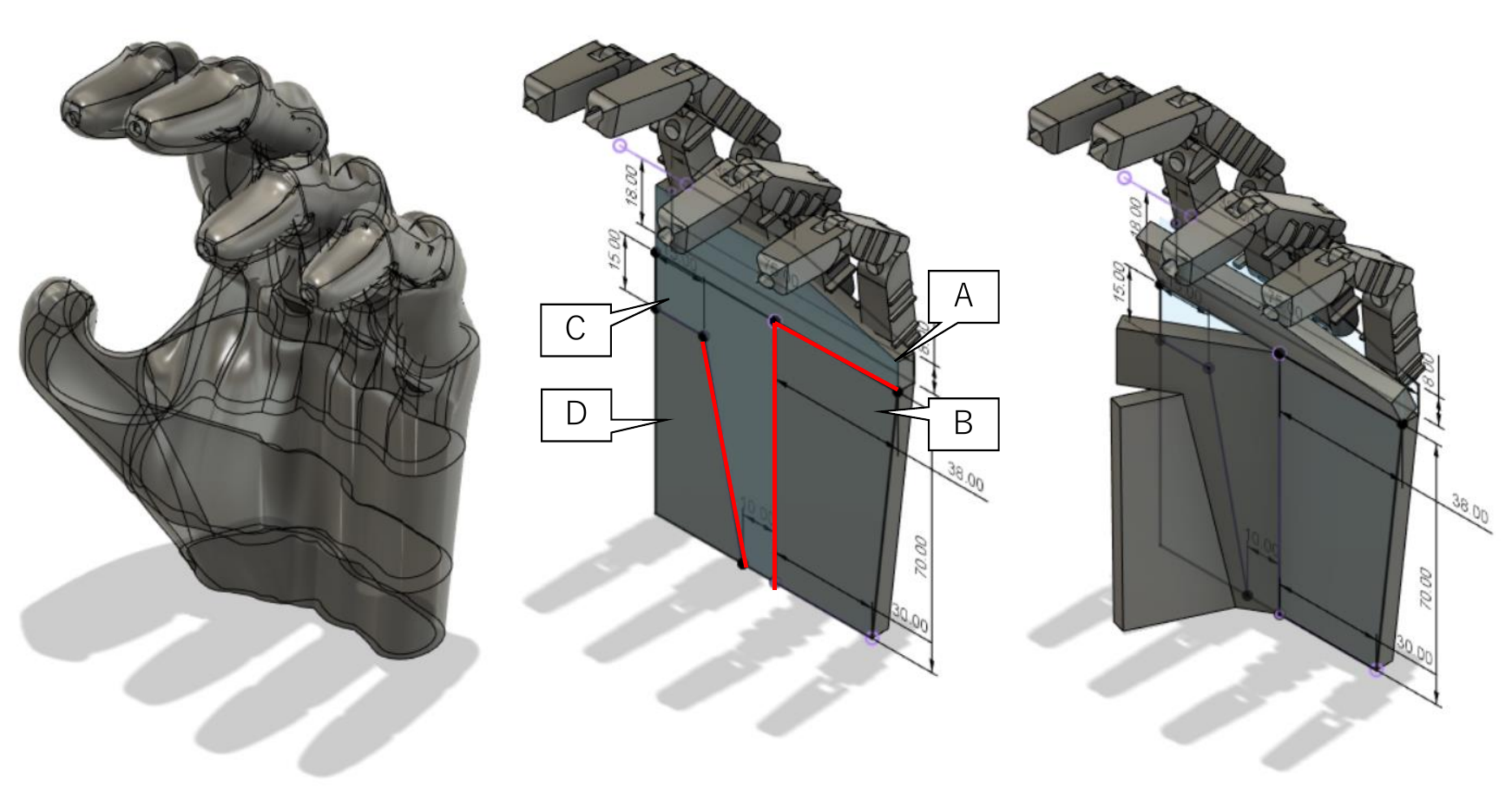}
  \vspace{-3.0ex}
  \caption{Comparison of the objects used to design palm of the epidermis}
  \label{palm}
  \end{figure}

\switchlanguage%
{%
  As described in \secref{subsec:comparison}, the palm structure is divided into four parts by the MP joints, the thenar muscles, and the hypothenar muscles. Comparative images of the objects used in the design and the completed epidermal-skeletal integrated model are shown in \figref{palm}.
  It is assumed that object B is fixed and object D is the movement of the thumb. The red edges are joined to each object, and the thumb is designed to achieve opposing motion by adjusting the length of each edge while moving object D.\par
  The part where the objects cut each other is made to bend easily as designed by varying the thickness of the model. In particular, since the area where the thenar muscles exist in the human body (object D) is subject to large deformations, we made it easier to fold it during the opposing motion by designing it to be bulged. \par
}%
{%
  \secref{subsec:comparison}で述べたように、掌の構造はMP関節と母指球筋、小指球筋によって4つのパーツに分けられる。設計に用いたオブジェクトと完成した表皮-骨格一体モデルの比較画像を\figref{palm}に示す。オブジェクトBは固定されており、オブジェクトDが母指の動きとなると仮定している。赤線のエッジがそれぞれのオブジェクトに結合されており、オブジェクトDを動かしながら確変の長さを調整することで母指が対向動作を実現できるように設計した。
  オブジェクト同士の切れ目になる部分はモデルの厚みを変化させることによって、設計通りに曲がりやすくしている。特に人体では母指球筋が存在する付近(オブジェクトD)は大きな変形をするため、膨らんだ設計にすることによって対向動作の際に折り畳まれやすくした。\par
}%

\section{Design of Other Parts, and How to Assemble the Robot Hand} \label{sec:thumb}
\subsection{Design of thumb and metacarpals parts}
\begin{figure}[htb]
  \centering
  \includegraphics[width=0.6\columnwidth]{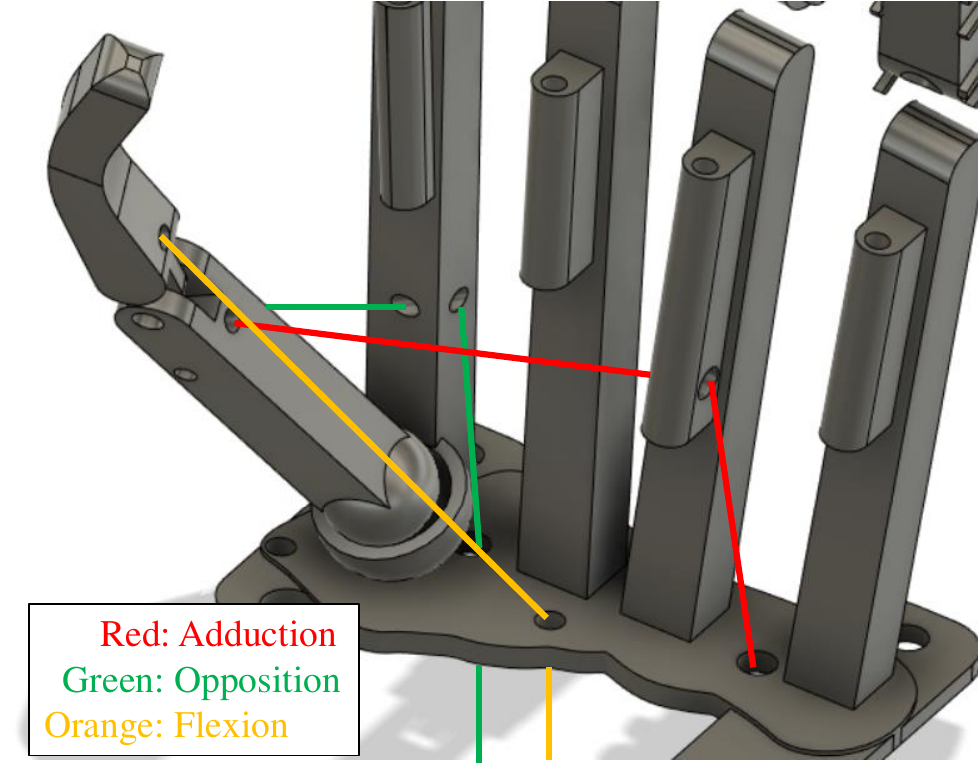}
  \caption{Wire paths for thumb and metacarpal parts}
  \label{thumb-wire}
  \end{figure}

\switchlanguage%
{%
In this section, the remaining components of the skin-skeleton integrated robot hand, the thumb and metacarpal parts, are described by comparing them with the human skeleton and tendon arrangement. The human thumb has three bones: the distal phalanx, proximal phalanx, and metacarpal bones, while the robot hand consists of two parts: a component joining the distal phalanx and proximal phalanx and a metacarpal bone.
Although the movement of the distal phalanx contributes to the grasping and hooking of small objects, it does not contribute to the realization of various grasping postures or the grasping of large objects, so it was omitted. The CM joint with two degrees of freedom, which is a characteristic of the thumb, is a spherical joint together with the metacarpal parts.\par
The wire path of the thumb is shown in \figref{thumb-wire}. As mentioned in \secref{subsec:comparison}, of the seven muscles used to drive the thumb, only the flexor pollicis brevis muscle, abductor pollicis muscles, and opponens pollicis muscle are reproduced.
The first is the flexor pollicis brevis muscle, which is the muscle used to flex the MP joint and is connected from the thumb basal bone to the flexor digitorum branch located in the wrist area.
The second is the adductor pollicis brevis muscle, which contributes to the adduction movement that fans out from the proximal phalanx to the metacarpals of the metacarpals of the metacarpals of the thumb. The second is the adductor pollicis brevis muscle, which contributes to the adduction movement that fans out from the proximal phalanx to the metacarpals of the metacarpals of the metacarpals of the thumb. 
The last, the opponens pollicis muscle, connects the metacarpal to the palmar carpal ligament, with the wires placed along a similar pathway. \par
The metacarpal bone is responsible for bringing together the tendons from the four fingers of the epidermal-skeletal integrated part and passing them to the wrist side. It is also made of ABS filament, which does not deform when tension is applied and improves the rigidity of the entire hand.
}%
{%
  本節では表皮-骨格一体型ロボットハンドを構成する残りの要素である親指,中手骨パーツについて、人間の骨格と腱配置との比較を行いながら述べる。人間の親指は末節骨、基節骨、中手骨の3本の骨を有しているが、本ロボットハンドでは末節骨と基節骨を結合した部品と中手骨の2つの部品によって成り立ってる。末節骨の動作は小さな物体の把持や引っ掛ける動作に寄与するものの、多様な把持姿勢の実現や大きな物体の把持には寄与しないため省略した。親指の特徴である2自由度を持つCM関節は、中手骨パーツと合わせて球状関節としている。\par
  親指のワイヤ経路を\figref{thumb-wire}に示す。\secref{subsec:comparison}でも述べたように、親指の駆動に用いられる7つの筋の中で短母指屈筋、母指内転筋、母指対立筋のみを再現した。1つ目の短母指屈筋はMP関節の屈曲を行うための筋であり、親指の基底骨から手首の種子骨周辺に存在する屈筋支帯へと繋がっているものであるため、同様に手首中央付近から引き出すワイヤ経路としている。2つ目の母指内転筋は基底骨から中指の中手骨まで扇状に広がる内転動作に寄与する筋肉であり、本ロボットハンドでは親指の中手骨から中指の中手骨までを繋ぐように配置している。最後の母指対立筋は中手骨から屈筋支帯までをつなげるもので、同様の経路でワイヤを配置した。\par
  中手骨は、表皮-骨格一体パーツの4指から出た腱をまとめて手首側に通す役割を持っている。また、ABS樹脂で制作されているため張力を加えた際に変形せず、ハンド全体の剛性を向上させている。
}

  \begin{figure*}[htb]
    \centering
    \includegraphics[width=1.9\columnwidth]{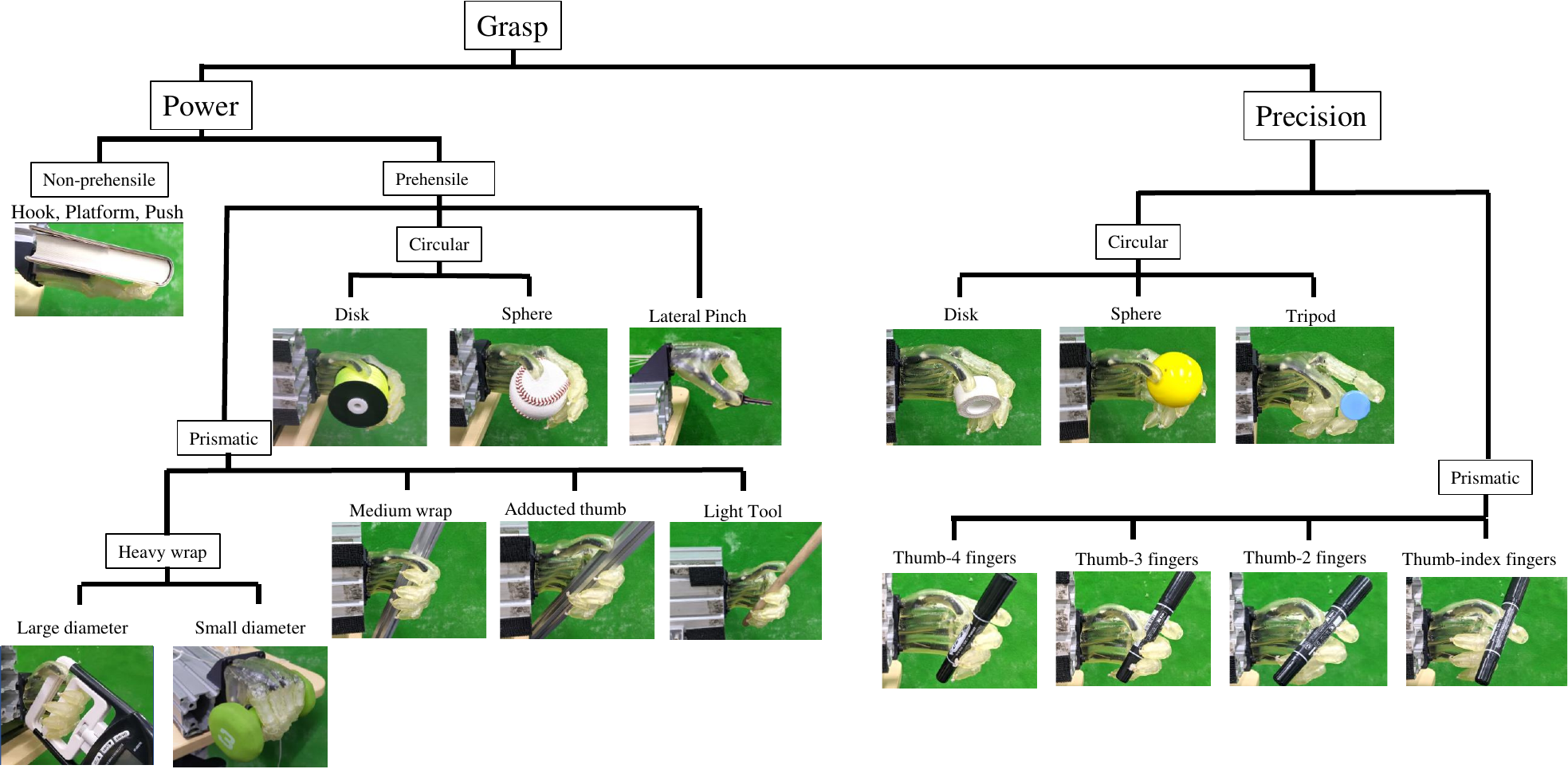}
    \caption{Classification of grasping methods according to \cite{34763} and results of achieved grasps.}
    \label{classify}
    \end{figure*}

\subsection{How to Assemble the Robot Hand}
\begin{figure}[htb]
    \centering
    \includegraphics[width=0.8\columnwidth]{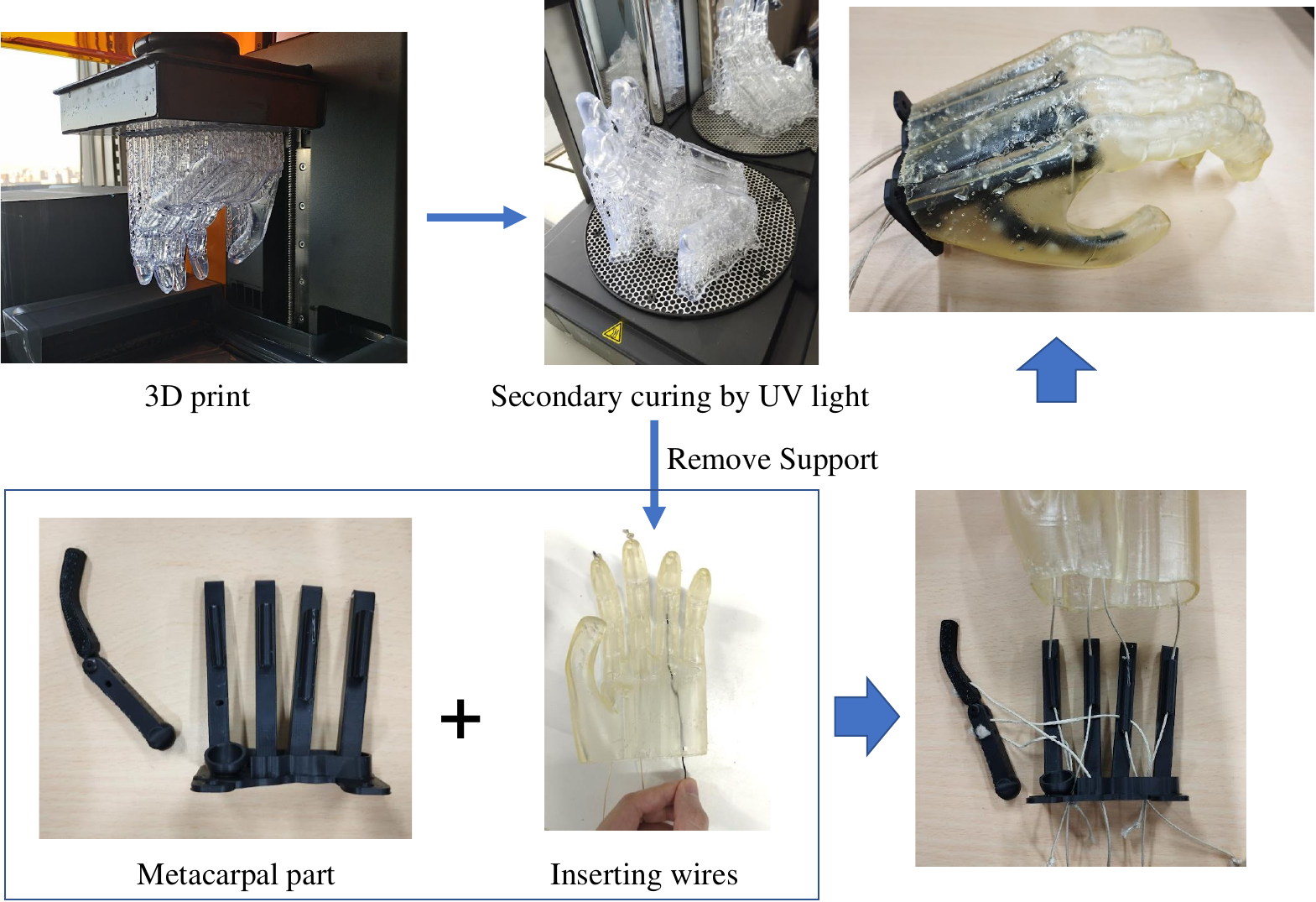}
    \vspace{-1.0ex}
    \caption{Parts that compose a robot hand and assembly method}
    \label{const}
    \end{figure}

\switchlanguage%
  {%
  This section describes the assembly method of the skin-skeleton integrated robot hand. This robot hand consists of only four parts (epidermal-skeletal integrated parts, metacarpal parts, basal and metacarpal bones of the thumb) and seven wires\figref{const}.
  The epidermal parts are designed using the aforementioned method and consist of the skeleton excluding the metacarpals from the index finger to the little finger, plus the epidermis covering them. 3D printed parts are completed by washing and curing, and are designed to be more tear-resistant by curing for 60 minutes, which exceeds the recommended effect time of 25 minutes.
  In passing the wires through the holes inside the skeleton of those surface skin parts, problems occurred, such as the wires stopping inside the holes due to friction of the surface skin material, or the skeleton part breaking when pushed in.
  Therefore, we have prepared a wire bonded to a metal wire, and succeeded in simplifying the wire placement by threading the metal wire first. After penetration, the adhesive part of the wire and metal wire is cut off, and the ball-knotted piece is fixed with a UV-curable adhesive at the fingertip hole. \par
  Parts other than the skin-skeleton integrated parts are fabricated by 3D printing with ABS filament. 
  First, a thumb part with a wire and shaft attached is attached to the thumb part of the skin part. Then, seven wires from the thumb and four other fingers are threaded through the metacarpal part, placed inside the skin part, and fixed with screws to complete. The entire assembly time is less than one hour, indicating that it is extremely simple to assemble.
  }%
  {%
  表皮骨格一体型ロボットハンドの組み立て方法について述べる。本ロボットハンドは表皮-骨格一体パーツ、中手骨パーツ、親指の基底骨と中手骨の4つの部品、7本のワイヤのみで構成されている\figref{const}。\\
  表皮パーツは前述の手法で設計され、人差し指から小指までの中手骨を除く骨格に加えそれらを覆う表皮を有している。3Dプリントしたものを洗浄、硬化させることで完成するが、推奨の効果時間である25分を超過する60分の硬化を行うことでより破れにくい設計にしている。それら表皮パーツの骨格内部の穴にワイヤを通す際、表皮の素材の摩擦によって穴の内部で止まってしまったり、押し込む際に骨格部が破断してしまうといった問題が発生した。そこでワイヤを金属線に接着したものを用意し、金属線を先に通すことでワイヤの配置を簡単にすることに成功した。貫通後はワイヤと金属線の接着部を切断し、玉結びしたものを指先の穴の部分で紫外線硬化性の接着剤で固定した。\\
  表皮以外のパーツはABS樹脂による3Dプリントによって製作される。まず、ワイヤと軸を取り付けた親指のパーツを表皮パーツの親指部分に取り付ける。その後、中手骨パーツに親指とその他4本の指から出る7本のワイヤを通し、表皮パーツの中に入れてねじ固定することで完成する。全体を通しての組み立て時間は1時間以下であり、極めて簡易に組み立てが可能であることがわかる。
  }

\section{Experiments} \label{sec:experiments}
\subsection{Items Grasping Task} \label{subsec:grasp-ex}
\switchlanguage%
{%
Object grasping experiments were conducted to verify the basic performance of this robot hand. The classification of grasping forms based on Cutkosky's classification of human grasping behavior\cite{34763} is shown in \figref{classify}.
This classification is divided into Power Grasp, which uses finger bellies and palms and Precision Grasp, which uses the tips of fingers for grasping, and is further subdivided according to differences in grasp shape based on the shape of the object. \par

In this experiment, the muscle module \cite{asano2015sensordriver} was connected to the seven tendons from the hand, and the muscle length was changed by an angular command while the object was fixed at a graspable point.
The results of the grasping experiments are also shown in \figref{classify}, and simple grasping of items such as Disc and Sphere was achieved with both the Power Grasp and Precision Grasp. \par

The detailed results of Power Grasp are described: Platform was achieved by placing a book on the palm of the hand without using actuators; Lateral Pinch was achieved by driving an actuator to hold the four fingers in a clenched position and to oppose the thumb.
Adductive thumb was achieved by forcing the thumb into a fixed position and flexing the other four fingers, although this posture cannot be met by control due to the absence of an actuator for the thumb's return direction.
The Light Tool was able to grasp a drumstick of 14[mm] in diameter, confirming that a highly accurate grip was achieved.
With the Heavy Grasp, we used a grip strength tester and a dumbbell. The result of the grip strength meter was 1.4[kg], which was not sufficient. This is thought to be because the subject was not able to clench his/her palm as well as humans do, and instead clenched the thumb and the other four fingers together.
On the other hand, as for the dumbbell, it was able to grasp 3 pounds when the palm was placed parallel to the floor surface and 8 pounds when the palm was placed perpendicular to the floor surface, indicating that it has high grasping performance for small diameter objects. \par
Precision Grasp was able to grasp a wide variety of objects even at the fingertips because the finger bellies were deformable. In addition, each finger could be operated to realize four different grasping motions using the opposed thumb.
}%
{%
  本ロボットハンドの基礎性能検証のため、物体把持実験を行った。\figref{classify}にCutkoskyによる人間の把持動作分類\cite{34763}を元にした把持形態の分類を示す。本分類は指の先端で把持をするPrecision Graspと指の腹や掌を用いるPower Graspに大別されており、その中でも物体の形状に基づく把持形状の違いによってより細かく分けられているものである。\par
  本実験ではハンドから出る7つの腱に筋モジュール\cite{asano2015sensordriver}を接続し、物体を把持可能な箇所に固定した状態で角度指令によって筋長を変化させた。把持実験の結果も\figref{classify}に示している。Power Grasp、Precision Graspの両者でDisc, Sphereといったアイテムの単純把持を達成した。\par
  Power Graspの詳細な結果について述べる。Platformはアクチュエータを用いておらず、ハンドの掌に本を乗せることで達成した。Lateral Pinchは、4指を握り込んだ状態にし母指を対立するためのアクチュエータを駆動することで達成した。Adductive thumbは、親指の復位方向のアクチュエーターが存在しないため制御によってこの姿勢を満たすことはできないが、親指を強制的に固定した状態で他の4指を屈曲させることで達成した。Light Toolでは直径14[mm]のドラムスティックを把持することができ、精度の高い握り込みが達成できていることが確認できた。
  Heavy Graspでは握力計とダンベルを把持させた。握力計の計測結果は1.4[kg]であり、十分とは言えなかった。これは人間のように掌との挟み込みが良好に行えず、親指と他4指の挟み込みになっていたからだと考えられる。一方でダンベルに関しては、掌を床面と平行に配置した際には3ポンド、床面と垂直に配置した場合は8ポンドを把持することができたことから、小径の物体に対しては高い把持性能があることがわかった。\par
  Precision Graspでは、指の腹が変形することから指先でも多様な物体の把持が可能だった。また、それぞれの指に自由度があることによって、opposed thumbを用いた4種類の把持動作を実現することができた。
}%

\subsection{Dial Operation Experiment}
\begin{figure}[htb]
  \centering
  \includegraphics[width=0.8\columnwidth]{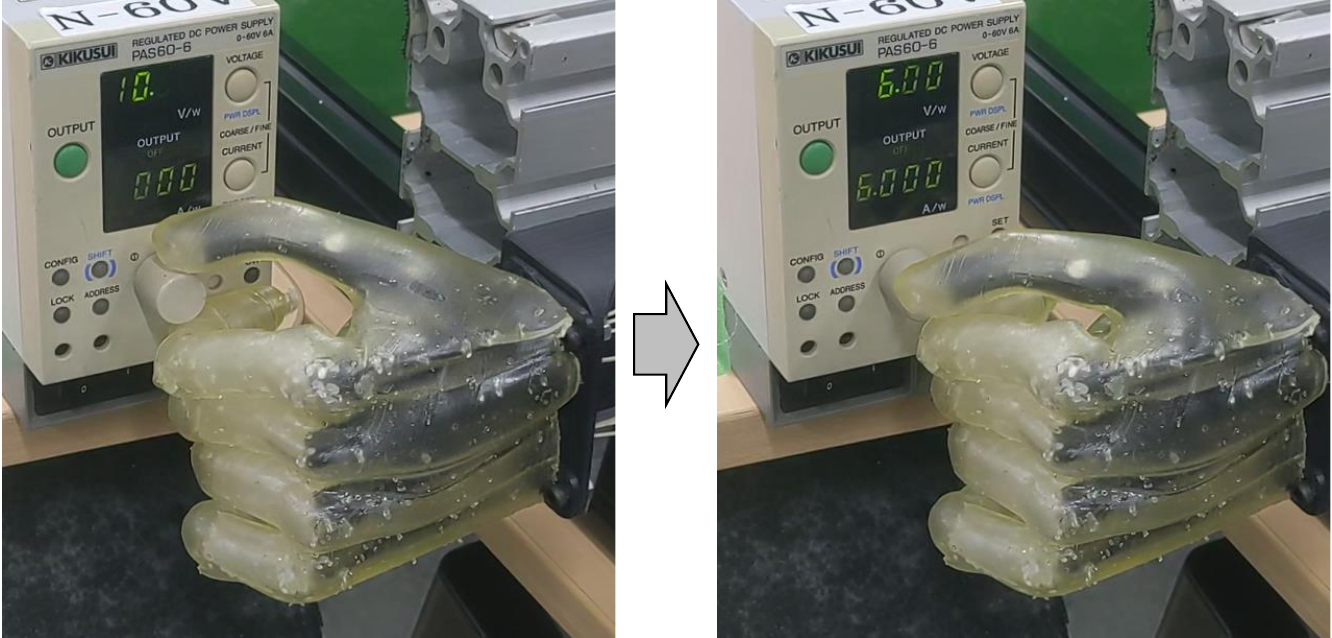}
  \vspace{-1.0ex}
  \caption{Operation of dials with thumb and index finger}
  \label{dial}
  \end{figure}

\switchlanguage%
{%
In order to verify object manipulation by driving the thumb, an experiment to manipulate a dial was conducted. The experiment was started with the dial held between the thumb and the index finger, and the objective was to rotate the dial in a semi-clockwise direction by simultaneously performing adduction and opposition movements of the thumb and flexion movements of the index finger. 
In the experiment, the dial was rotated by pulling the tendon 9[mm] in thumb adduction, 9[mm] in thumb opposition, and 4[mm] in index finger flexion. The dials were from a power supply unit that was not connected to anything, and were tested to see if they varied from a display of 10[V] in the mode of varying voltage. \par

}%
{%
  母指の駆動による物体操作に関する検証を行うため、ダイアルの操作実験を行った。母指と示指でダイアルを挟み込んだ状態で実験を開始し、母指の内転、対立動作と示指の屈曲動作を同時に行うことによってダイアルを半時計回りの回転させることを目的とする。実験では母指内転の筋長を9[mm],母指対立の筋長を9[mm], 示指屈曲の筋長を4[mm]引くことによって、ダイヤルの回転を行った。ダイヤルは何とも接続されていない電源装置のものを用い、電圧を変化させるモードの中で10[V]の表示から変化するかを確かめた。\par
  ダイアルの回転によって、10[V]から6[V]まで電源装置の値が変化することを確認できた(\figref{dial})。これにより、母指の内転-対立のアクチュエータが有効に動作していることを確認することができたとともに、ネジ回しなどの複雑な動作に拡張することが期待される。 
}%

\subsection{Impact Resistance Experiment}
\begin{figure}[htb]
  \centering
  \includegraphics[width=1.0\columnwidth]{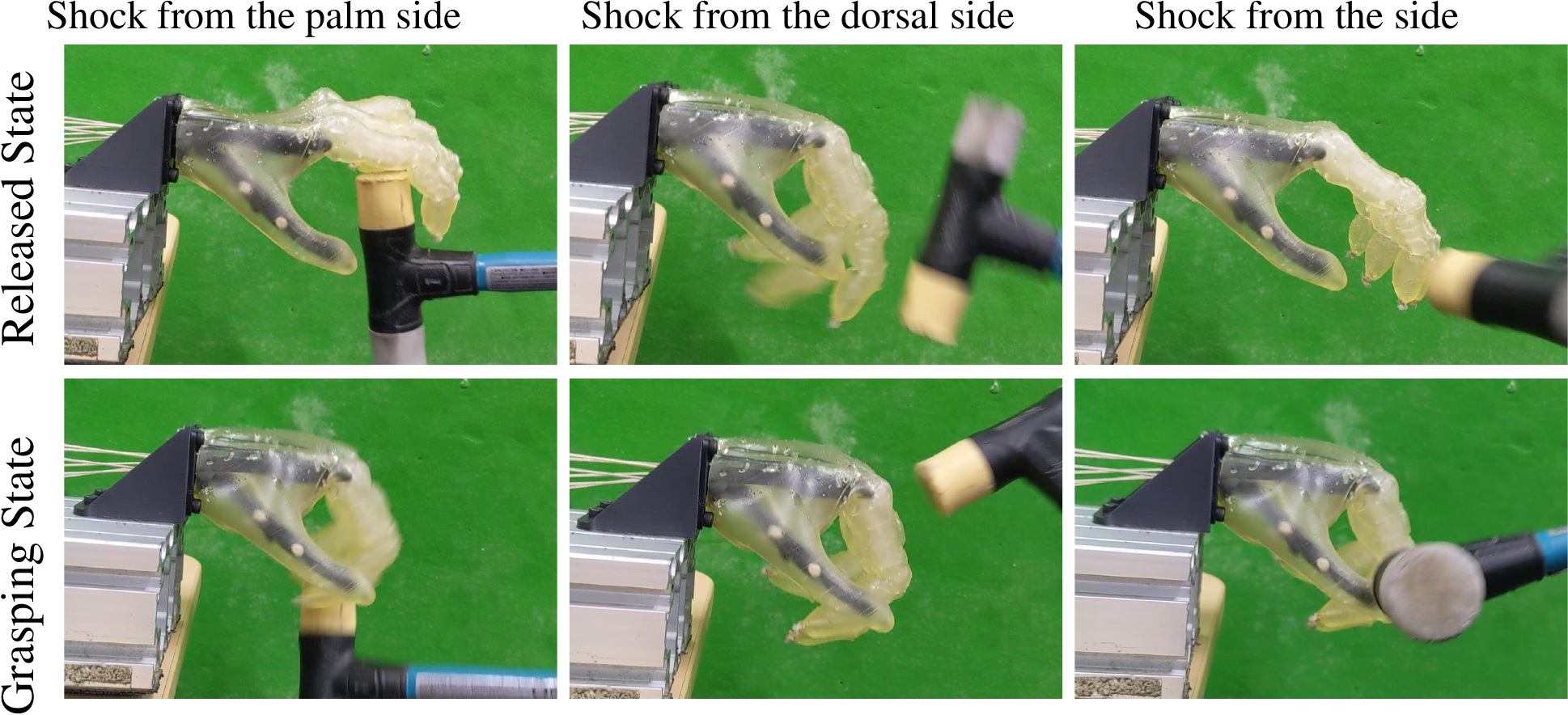}
  \vspace{-3.0ex}
  \caption{Result of shock resistance experiment}
  \label{shock}
  \end{figure}

\switchlanguage%
{%
To verify the adaptability of the robot hand to the environment, a hammer impact was applied to the hand with and without tension. The direction of impact was palm, dorsal, and lateral. In all six types of experiments, the hammers passed through the impacts without any problem and returned to their original positions within 1[s]. The results of the experiments are shown in \figref{shock}.
}%
{%
  環境適応性の検証のため、張力を加えている場合と加えていない場合にハンマーで衝撃を加えた. 衝撃を加える向きは掌側,背側,側面の3種である. いずれの6種類の実験においても問題なく衝撃を受け流し,1[s] 以内に元の位置に戻った. 実験の結果を\figref{shock}に示す.
}%

\section{Conclusion} \label{sec:conclusion}
\switchlanguage%
{%
In this study, we proposed a skin-skeleton integrated robotic hand with the aim of fabricating a soft robotic hand that is both easy to assemble and complex to operate. While general robot hands have a problem that the number of parts increases with the number of degrees of freedom, this robot hand has 15 degrees of freedom and is composed of mostly single parts, which makes it easy to fabricate and assemble. \par
The most important element to realize a robot hand with multiple degrees of freedom with a single part is the skin-skeleton integrated part. This is a part that is a skin and skeleton formed at once from a flexible material, and it has no axes at all, yet it realizes 12 DOFs.
To achieve this, the joints have a dorsal crease and a hole in the palm, which allow the joints to flex with little tension. \par
The thumb part is driven by three muscles, which are designed to be restored using the elasticity of the skin to obtain 3-DOFs by adding flexion to the 2-DOF joint structure. These three muscles were designed to realize a human grasping posture by referring to the muscle arrangement of the human body.\par 
Grasping experiments were conducted to verify the performance of the fabricated skin-skeleton integrated robot hand. In the grasping experiment, all of the 16 grasping postures were achieved with human assistance, indicating the possibility of achieving movements similar to those of humans.
In addition to grasping, dialing was also achieved, indicating the possibility of extending tool manipulation by driving the thumb. \par
In this study, the thumb was driven by a simple muscle length command and was not controlled by the grasping state. In the future, we aim to obtain the status of each finger by placing sensors and performing other controls.
}%
{%
  本研究では組み立て性と動作の複雑性を両立するソフトロボットハンドを製作することを目的とし, 表皮-骨格一体型ロボットハンドを提案した. 本ロボットハンドは大部分が単一パーツのよって構成されていることから製作や組み立てが簡単である一方で、15自由度を持ち様々な把持形態を実現可能である。\par
  表皮-骨格一体パーツは関節に背側のシワと手の平側の穴を有しており、これらによって小さい張力で屈曲することが可能になっていることを確認した。また、人体の腱配置を模倣するために骨格内部にワイヤガイドを設計し、適切なモーメントアームを確保できることを数式により確認した。\par
  親指パーツは3筋を用いて駆動しており、皮膚の弾性を用いて復元するような設計にすることで2自由度の関節構造+屈曲の３自由度を得るための筋配置について述べた。また、中手骨パーツを含めた３パーツの製作方法と組立方法について述べ、表皮-骨格一体パーツを一体成型したことによる組み立ての簡易性を示した。\par
  製作した表皮-骨格一体ロボットハンドの性能検証のため、把持実験を行った。把持実験では人間の介助はあったものの16種の把持姿勢の全てを達成し、人体と同様の動作が実現出来る可能性を示した。把持以外にもダイヤル操作なども達成したことで、母指の駆動による道具操作の拡張性を示した。\par
  本研究では単純な筋長指令によって駆動させており、把持状態などによる制御は行えてはいなかった。今後はセンサを配置することで各指の状況を取得し他制御を行うことを目標とする。
}%

{
  \bibliographystyle{IEEEtran}
  \bibliography{main}
}

\end{document}